\documentclass[lettersize,journal]{IEEEtran}
\usepackage{amsmath,amsfonts}
\usepackage{algorithmic}
\usepackage{algorithm}
\usepackage{array}
\usepackage[caption=false,font=normalsize,labelfont=sf,textfont=sf]{subfig}
\usepackage{textcomp}
\usepackage[hidelinks]{hyperref}
\usepackage{stfloats}
\usepackage{enumerate}
\usepackage{url}
\usepackage{verbatim}
\usepackage{graphicx}
\usepackage{booktabs}
\usepackage{multirow}
\usepackage{mathbbol}
\usepackage{tabularx}
\usepackage{tabulary,xcolor}
\usepackage{cite}
\usepackage{soul}
\usepackage{pifont}
\newcommand{\cmark}{\ding{51}}%
\newcommand{\xmark}{\ding{55}}%
\usepackage[capitalize]{cleveref}
\crefname{section}{Section}{Sections}
\crefname{table}{Table}{Tables}

\begin{document}

\title{S3Former: Self-supervised High-resolution Transformer for Solar PV Profiling }


\author{Minh Tran~\IEEEmembership{Student,~IEEE}, Adrian De Luis,  Haitao Liao~\IEEEmembership{Member,~IEEE}, Ying Huang~\IEEEmembership{Member,~IEEE}, \\ Roy McCann~\IEEEmembership{Member,~IEEE}, Alan Mantooth~\IEEEmembership{Fellow,~IEEE}, Jack Cothren~\IEEEmembership{Member,~IEEE}, Ngan Le~\IEEEmembership{Member,~IEEE}
}

\markboth{Journal of \LaTeX\ Class Files,~Vol.~14, No.~8, August~2021}%
{Shell \MakeLowercase{\textit{et al.}}: A Sample Article Using IEEEtran.cls for IEEE Journals}


\maketitle

\begin{abstract}
As the negative impact of climate change escalates, the global necessity to transition to sustainable energy sources becomes increasingly evident. Renewable energies have emerged as a viable solution for users, with Photovoltaic (PV) technology being a favored choice for small installations due to its high reliability, competitive market and increasing efficiency. Accurate mapping of PV installations is crucial for understanding the trend of technology adoption and informing energy policy-making. To meet this need, S3Former is introduced, which is designed to segment solar panels from aerial imagery and provide size and location information critical for analyzing the impact of such installations on the grid. Although computer vision has become a preferred choice for such implementations, solar panel identification is challenging due to factors such as time-varying weather conditions, different roof characteristics,  Ground Sampling Distance (GSD) variations and lack of appropriate initialization weights for optimized training. To tackle these complexities, S3Former features a Masked Attention Mask Transformer incorporating a self-supervised learning pretrained backbone. Specifically, the model leverages low-level and high-level features extracted from the backbone and incorporates an instance query mechanism incorporated on the Transformer architecture to enhance the localization of solar PV installations. Moreover, a self-supervised learning (SSL) phase (pretext task) is introduced to fine-tune the initialization weights on the backbone of S3Former, leading to a noticeable improvement on the results. To rigorously evaluate the performance of S3Former, diverse datasets are utilized, including GGE (France), IGN (France), and USGS (California, USA), across different GSDs. Our extensive experiments consistently demonstrate that the proposed model either matches or surpasses state-of-the-art models (SOTA) and validate the benefit of using the SSL method to improve the segmentation architecture. Source code is available upon acceptance.


\end{abstract}


\section{Introduction}

Photovoltaic (PV) energy production has grown rapidly in recent years due to improved cell manufacturing technology \cite{PETERS20192732}, competitive market, and increasing energy costs \cite{LEMAY2023113571}. In 2022, the U.S. Energy Information Administration reported a 440\% increase in small-scale PV installation capacity, reaching 39.5 GW from 7.3 GW in 2014 \cite{EIA2022}.

While solar panels offer a sustainable source of energy, they are faced with challenges for energy forecasting \cite{PIERRO2022983}. PV energy relies on weather conditions and physical characteristics of the PV module to operate. Although this information can be leveraged to calculate the predicted output of solar farms over a time period, small-scale PV operates in decentralized systems, and the real-time energy output of these systems is usually invisible to Transmission System Operators \cite{7347457}. Therefore, accurate energy forecasting of such systems is crucial, especially in those areas with high PV adoptions, to prevent grid congestion and imbalances \cite{PIERRO2022983}.

To gain real-time insights into PV distribution and energy production, recent research has focused on profiling PV installations using remote sensing imagery from unmanned aerial vehicles and satellites \cite{YU20182605}, coupled with Machine Learning and Computer Vision. Remote Sensing (RS) images provide an overhead view of the Earth's surface and define the level of detail as Ground Sample Distance (GSD). RS imagery serves multiple practical purposes, including climate change analysis \cite{ONEILL2013413}, natural disaster monitoring \cite{DBLP:journals/corr/abs-1901-04277}, urban planning \cite{9324098}\cite{Effat2013AMA}, and land cover analysis \cite{NEURIPS_DATASETS_AND_BENCHMARKS2021_4e732ced}.

Deep Learning (DL) has shown remarkable image understanding capabilities, even surpassing human performance. It has been applied to extract solar PV distribution through a two-step process: a classifier is trained to identify solar panels to then be segmented such as Kasmi, et al., \cite{kasmi2023unsupervised}, HyperionSolarNet\cite{parhar2022hyperionsolarnet}. However, existing DL-based methods have three major limitations:

\begin{enumerate}[(i)]
  \item The existing works employ a two-stage framework, consisting of separate classifier and segmentor networks. This approach heavily relies on the performance of classifier network, leading to suboptimal learning in the segmentation network. They overlook the unique challenges posed by solar panels, in which solar cells can be arranged on multitude of non-repeating structures. This challenge is known as \textit{intra-class homogeneity} and can hinder network performance as it may lead to missing elements from the resulting mask. This problem is further exacerbated by a variety of choices for cell materials.
  \item The specific characteristics of aerial imagery are not properly captured by mainstream initialization weights, with models missing detailing present on an image. This problem is exacerbated at higher GSDs, significantly impairing the ability of DL methods to detect \textit{small objects} and potentially causing the resultant masks to omit portions of solar panels.
  \item These existing DL frameworks are originally designed for natural images and do not adequately address the specific challenges encountered in solar PV imagery, leading to the misidentification of solar cells. This confusion is due to solar arrays sharing similar characteristics, such as shape or color, with commonly occurring objects (e.g., windows, roof tiles). This challenge is known as \textit{inter-class heterogeneity}.
\end{enumerate}

\begin{figure*}[!t]
\centering
\includegraphics[width=\linewidth]{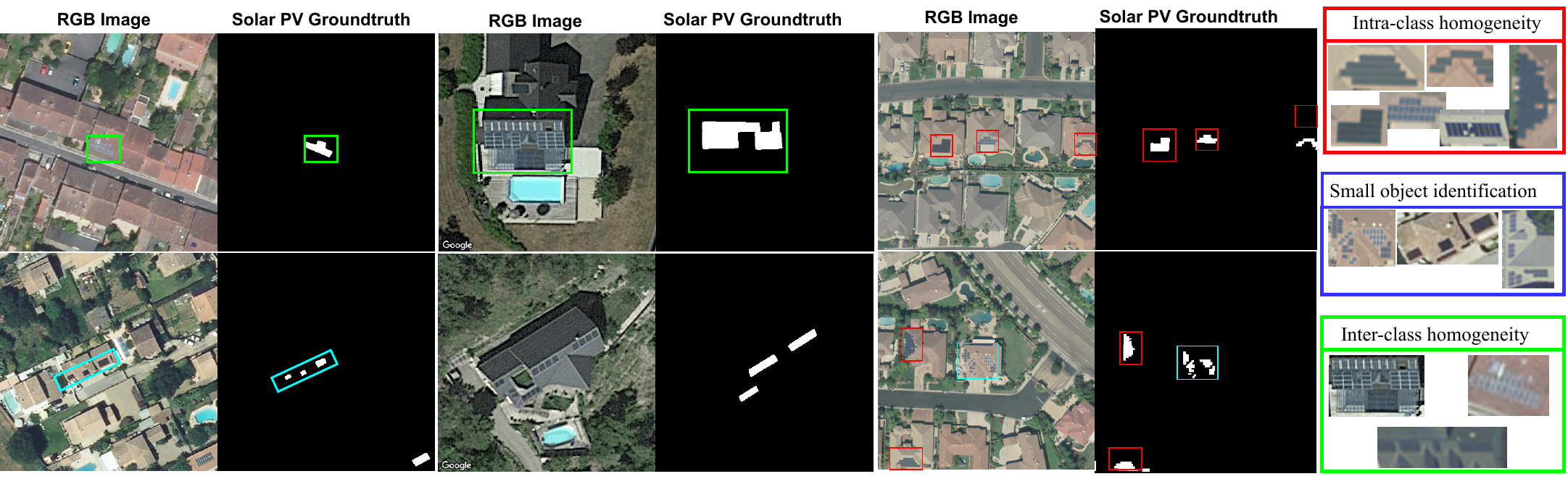}
\caption{Examples of challenging characteristics of solar PV segmentation on (a) IGN France, (b) GGE France, (c) USGS California.. Within a class, there is a large diversity in appearance: intra-class heterogeneity (red), some different classes share the similar appearance: inter-class homogeneity (green), solar PV are dense and small such that they are hardly identifiable (blue).}
\label{fig:solar_pv_challenges}
\end{figure*}

Problems (i), (ii), (iii) are depicted in Fig. \ref{fig:solar_pv_challenges}. 
To overcome these limitations, we introduce a \textbf{S3Former}, an \textbf{end-to-end transformer-based segmentation network},
featuring a Masked
Attention Mask Transformer incorporating a self-supervised
learning pretrained backbone for feature extraction.

Specifically, to address issue (i), we develop S3Former, a novel end-to-end transformer-based segmentation network. Unlike previous approaches that require localization prior to segmentation, S3Former seamlessly segments solar panels directly from images. This approach eliminates suboptimal learning segmentation networks.
Additionally, leveraging transformer-based learning allows us to effectively address issue (ii). By leveraging  attention mechanisms, the models are able to learn correlations among multi-scale image features from low to high resolutions. Consequently, we achieve improved segmentation accuracy, particularly in the identification of small objects.
Regarding issue (iii), prior research employs backbone models pretrained on common scenes from large image datasets, resulting in visual features lacking exposure to representations found in aerial imagery. This deficiency is critical for accurate solar panel segmentation. Therefore, we introduce a pretraining phase to our network using self-supervised learning, also known as a pretext task. This approach allows us to leverage numerous unannotated aerial images to enhance our model's understanding of aerial imagery before tackling the segmentation task (downstream task). The core concept of this pretext task is to eliminate non-semantic features from image representations. In essence, the representation of an augmented view of an image should predict the representation of another augmented view of the same image.

In summary, the contributions of this work are as follows:

\begin{itemize}
    \item We propose S3Former, a novel end-to-end transformer-based segmentation framework for solar panels segmentation. The attention mechanism in our transformer modules learn correlations among multi-scale image features from low to high resolutions, lead to accuracy improvement, particularly in the identification of small objects.
    \item We introduce a pretraining phase to our network using self-supervised learning that allows us to leverage numerous unannotated aerial images to enhance our model's understanding of aerial imagery before tackling the downstream task.
    \item We conduct extensive experiments on three publicly available datasets (i.e. USGS, California~\cite{California2016}; IGN, France~\cite{Kasmi_2023}; USGS, France~\cite{Kasmi_2023}), comparing the performance of S3Former to the current SOTA methods.
\end{itemize}

    


    

\section{Related Work}

\subsection{Deep Learning-based Solar PV Analysis} 
In recent years, high-fidelity solar mapping has gained significant importance due to the widespread adoption of photovoltaic (PV) energy, improved aerial imagery resolution, and advancements in deep learning (DL) techniques \cite{MALOF2016229,8127092}. Early efforts focused on binary image classification to detect PV installations.
DeepSolar \cite{YU20182605} introduced a dual-stage Convolutional Neural Network (CNN) for classification and segmentation, marking a significant milestone. Subsequent work integrated state-of-the-art CNN methods into classifier and segmentation architectures, while others concentrated solely on segmentation \cite{ZHUANG2020106283,9300636}. These methods also enable PV capacity estimation \cite{MAYER2022118469,kasmi2023unsupervised} and socioeconomic analyses \cite{YU20182605}.
A major challenge lies in creating suitable datasets, requiring time-consuming annotation and addressing data sensitivity concerns. Organizations like IGN and USGS offer freely accessible aerial images, but researchers often face limitations in annotation and data release \cite{parhar2022hyperionsolarnet,Kasmi_2023,California2016}.

\subsection{Transformer}
Transformer \cite{vaswani2017attention} and Vision Transformer (ViT) \cite{dosovitskiy2020image} have drawn significant interest in recent research. Initially developed for Natural Language Processing (NLP), the Transformer \cite{vaswani2017attention} employs a self-attention mechanism to improve learning of long-range dependencies within data, facilitating parallelization for faster training on large datasets across multiple nodes. These attributes have proved invaluable for various NLP tasks \cite{9222960}. ViT divides 2D images into patches, treating them akin to words in an NLP Transformer. This approach has shown efficacy, particularly with large datasets, enabling effective information aggregation from lower levels to produce superior global context compared to CNNs. ViT has been successfully applied to tasks such as image recognition \cite{touvron2021training}, video captioning \cite{yamazaki2022vlcap} \cite{yamazaki2023vltint}, action localization \cite{vo2022contextual, vo2023aoe}, aerial imaging \cite{kasmi2023unsupervised}, object detection \cite{sun2021sparse, tran2022aisformer}, image segmentation \cite{tran2022aisformer}, and medical imaging \cite{nguyen2023embryosformer}, showcasing its ability to synthesize global information.
Most deep learning (DL) methods for solar PV analysis are currently adaptations of existing DL techniques customized for solar PV data. However, these approaches often struggle with the unique challenges of solar PV datasets, such as low resolution, intra-class variability, and inter-class overlap. S3Former, on the other hand, is tailored specifically for extracting solar PV data while effectively addressing these challenges. 

\subsection{Self-supervised Learning}
Self-supervised learning (SSL) is a rapidly growing technique for learning feature representations without labeled data, as opposed to supervised pre-training. Much of the research in SSL focuses on discriminative techniques termed \emph{instance classification} \cite{chen2020simple,dosovitskiy2016discriminative,he2020momentum,wu2018unsupervised}. In this approach, each image is treated as a separate class, and the model learns to differentiate between them using various data augmentations. However, employing an explicit classifier for all images encounters scalability issues as the dataset size increases \cite{dosovitskiy2016discriminative}. To overcome this, Wu \textit{et al.} \cite{wu2018unsupervised} propose using a noise contrastive estimator (NCE) \cite{gutmann2010noise} for instance comparison instead of classification. Nevertheless, this method requires comparing features across a large number of images simultaneously, often necessitating large batches \cite{chen2020simple} or memory banks \cite{he2020momentum, wu2018unsupervised}. Various adaptations have emerged, such as automatic instance grouping through clustering techniques \cite{asano2019self,caron2019unsupervised}.

Recent studies have demonstrated the feasibility of learning unsupervised features without explicitly discriminating between images. This is typically achieved by first training the network unsupervised with a pretext task in the pre-training stage, then finetuning it on a downstream task. Grill et al. \cite{grill2020bootstrap} introduce BYOL, a metric-learning framework where features are trained by aligning them with representations from a momentum encoder. While methods like BYOL can function without a momentum encoder, they may experience decreased performance \cite{chen2020exploring,grill2020bootstrap}. This observation is supported by several other approaches, which demonstrate the alignment of representations, training features to match a uniform distribution, or utilizing whitening techniques \cite{gidaris2020learning,zbontar2021barlow}. Inspired by BYOL, DINO \cite{caron2021emerging} adopts a similar architecture but introduces the similarity matching loss, along with self-distillation \cite{tarvainen2017mean}.

\begin{figure}[!t]
    \centering
    \includegraphics[width=\linewidth]{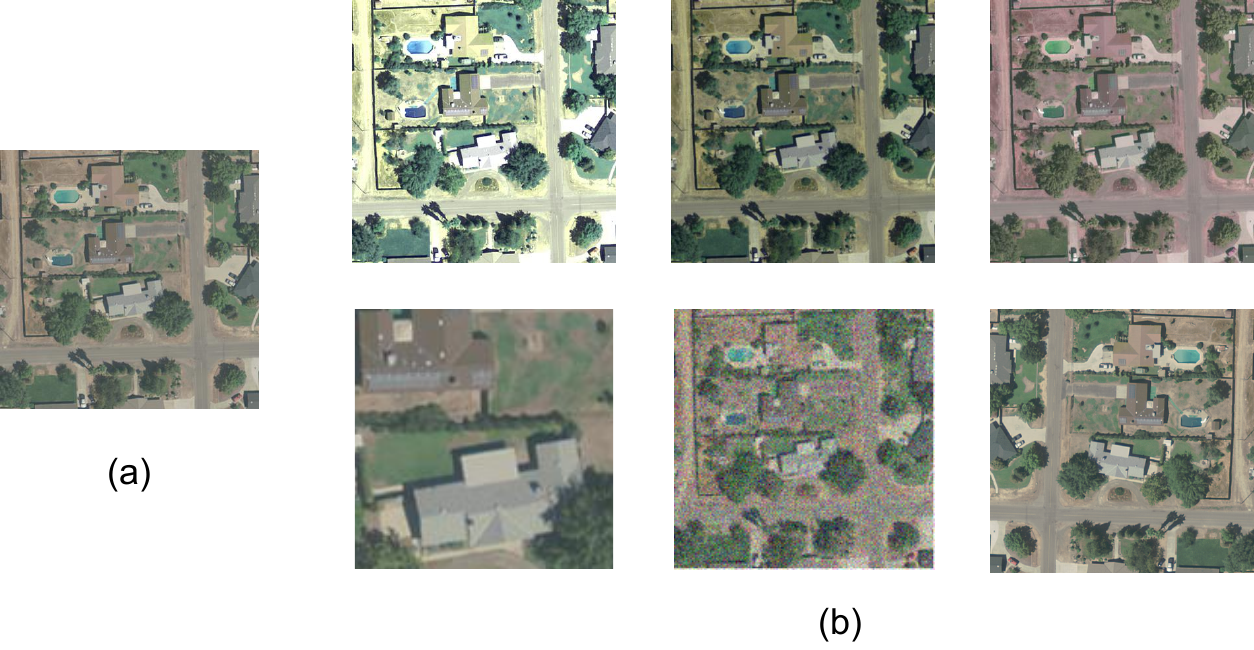}
    \caption{Examples of six augmentations for the pretext task self-supervised learning. (a): original image. (b) top line: color jitter; bottom line, from left to right: random cropping, Gausian noise, horizontal flip.}
    \label{fig:augmentations}
\end{figure}

\begin{figure*}[!t]
\centering
\includegraphics[width=0.85\textwidth]{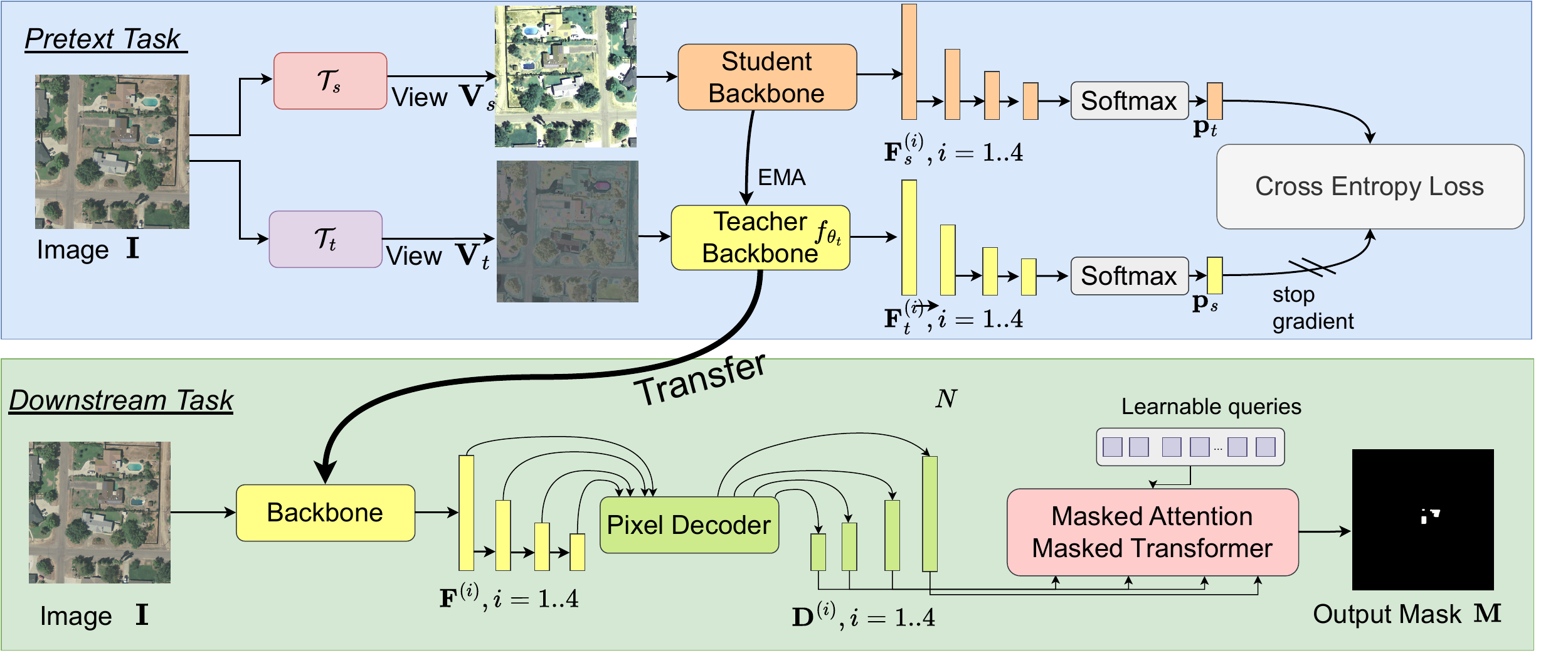}
\caption{Overall pipeline of the proposed S3Former. The training pipeline includes two phases: Pretext task and Downstream task. The goal of the pretext task is to learn the optimal parameter such that the backbone network can extract similar representations from both $\mathbf{V}_s$ and $\mathbf{V}_t$, regardless of non-semantic factors introduced by augmentation. In the downstream task, the Pixel Decoder learns the correlations within multi-scale feature maps and utilizes them to decode enriched feature maps. Following this, the Masked Attention Mask Transformer integrates these decoded feature maps with $N$ learnable queries, employing a masked attention mechanism to accurately predict the segmentation mask.}
\label{fig:network_arch}
\end{figure*}

\section{S3Former}
We 
design the architecture of our S3Former with three components: Backbone Network, Pixel Decoder, and Masked Attention Mask Transformer. More specifically:
\begin{itemize}
    \item \textbf{Backbone Network}: This module is for extracting visual features from the input image $\mathbf{I} \in \mathbb{R}^{H\times W\times 3}$. The extracted visual feature is a set of four multi-scale feature maps ${\mathbf{F}^{(i)}}$, where ${i=1..4}$. These feature maps are represented as $\mathbf{F}^{(1)} \in \mathbb{R}^{C_{\mathbf{F}^{(1)})}\times \frac{H}{4} \times \frac{W}{4}}$, $ \mathbf{F}^{(2)} \in \mathbb{R}^{C_{\mathbf{F}^{(2)}}\times \frac{H}{8} \times \frac{W}{8}}$, $ \mathbf{F}^{(3)} \in \mathbb{R}^{C_{\mathbf{F}^{(3)}}\times \frac{H}{16} \times \frac{W}{16}}$, and $ \mathbf{F}^{(4)} \in \mathbb{R}^{C_{\mathbf{F}^{(4)}}\times \frac{H}{32} \times \frac{W}{32}}$, where $C_{\mathbf{F}^{(1)}}, C_{\mathbf{F}^{(2)}}, C_{\mathbf{F}^{(3)}}, C_{\mathbf{F}^{(4)}}$ denote the number of channels.
    
    \item \textbf{Pixel Decoder} This module works as a visual feature enhancement with attention mechanism. The goal of the Pixel Decoder is to enhance the relationships between the multi-scale feature maps $\mathbf{F}^{(1)}, \mathbf{F}^{(2)}, \mathbf{F}^{(3)}, \mathbf{F}^{(4)}$ created by the backbone and learn their correlations. This produces richer encoded features in the pixel decoder layers $\mathbf{D}^{(1)}, \mathbf{D}^{(2)}, \mathbf{D}^{(3)}, \mathbf{D}^{(4)}$ with the same spatial feature shape as $\mathbf{F}^{(1)}, \mathbf{F}^{(2)}, \mathbf{F}^{(3)}, \mathbf{F}^{(4)}$, respectively. 

    \item \textbf{Masked Attention Mask Transformer} Given the enriched visual features extracted from the preceding two modules, denoted as $\mathbf{D}^{(1)}, \mathbf{D}^{(2)}, \mathbf{D}^{(3)}, \mathbf{D}^{(4)}$, this module focuses on predicting the segmentation mask $\mathbf{M} \in \mathbb{R}^{H\times W}$ for identifying solar panels within the image. Each pixel $\mathbf{M}_i$ is assigned a value of 0 if it corresponds to the background; otherwise, it is set to 1, indicating the presence of a solar panel.
    This module employs a transformer-based approach for mask prediction, integrating learnable queries to represent the instance mask within the image. The process involves decoding the learnable queries from the extracted visual features and associating them with the visual features to generate the segmentation mask.
\end{itemize}

In order to train our S3Former to perform the task, we do it in two phases: Pretext Task (\cref{sec:pretext}) and Downstream Task (\cref{sec:downstream}).
In the pretext task phase, we pretrain the Backbone Network with a pretext task in a self-supervised manner. This phase enhances the multi-scale features provided by the Backbone Network with stronger representations of aerial imagery. Thus, it enables us to address the aforementioned intra-class homogeneity issues effectively.
In the downstream phase, we begin by initializing the pretrained weights of the Backbone Network into our S3Former. We then proceed to train it on the segmentation task, supervised by the segmentation ground truth.

\subsection{Pretext Task}
\label{sec:pretext}
A \textbf{Backbone Network}, foundational for feature extraction, is pre-trained on various tasks, notably effective in computer vision, e.g., object detection. AlexNet~\cite{krizhevsky2017imagenet} and VGG~\cite{simonyan2014very} are early deep learning backbones, while ResNets~\cite{he2016deep}, like ResNet-18/34/50, are prominent for object detection. ResNets create multi-scale features, aiding computational tasks.

Typically, Backbone Networks are pretrained on large-scale general image datasets like ImageNet~\cite{russakovsky2015imagenet}, where classification tasks supervise the learning process. These datasets predominantly consist of images depicting common objects and scenes. Previous research in solar segmentation~\cite{de2023solarformer, 9300636, MAYER2022118469} often leverages pretrained Backbone Networks (e.g., ResNet~\cite{he2016deep}) that are pretrained on these datasets, followed by fine-tuning on the specific downstream task of solar panel segmentation.
However, one limitation of these pretrained networks is their lack of exposure to aerial imagery, which is crucial for tasks like solar panel detection. As conventional datasets seldom contain aerial images, there's a gap in representation. To address this, we employ SSL techniques to train the network without relying on labeled data. This approach enhances the network's ability to capture features specific to aerial imagery, thus bridging the gap in representation.

Many successful SSL approaches build upon the cross-view prediction framework introduced in~\cite{becker1992self}.
The idea is to remove non-semantic feature out of the image representation. That means
the representation of an augmented view of an image should be predictive of the representation of another augmented view of the same image.

Consider a backbone network, denoted as $f$ and parameterized by $\theta$. Let $\mathcal{T}_s$ and $\mathcal{T}_t$ represent augmentations sampled from a set including color jitter, random crop, Gaussian noise, and horizontal flip (as illustrated in \cref{fig:augmentations}). These transformations, when applied to an image $\mathbf{I}$, generate two augmented views, $\mathbf{V}_s$ and $\mathbf{V}_t$. The goal of the pretext task is to learn the optimal parameter set $\theta^{*}$ such that the network can extract similar representations from both $\mathbf{V}_s$ and $\mathbf{V}_t$, regardless of non-semantic factors introduced by augmentation.

In essence, the objective is for $f_{\theta}$ to produce comparable representations for diverse augmentations of the same image. However, directly minimizing the distribution discrepancy between the representations output by $f_\theta$ could lead to a problematic scenario known as collapse. This occurs when the network learns to output a constant representation, ignoring the input data, in order to minimize the difference between representations. 
To address this issue, we employ the self-distillation technique used in DINO~\cite{caron2021emerging}. This involves introducing both a student backbone, $f_{\theta_s}$, and a teacher backbone, $f_{\theta_t}$, which share similar architectures with $f$ but are parameterized by two different sets of weights, $\theta_s$ and $\theta_t$, respectively.
In traditional distillation methods~\cite{44873}, a student network, $f_{\theta_s}$, is trained to replicate the outputs of a given teacher network, $f_{\theta_t}$. However, in the context of SSL, this technique is termed self-distillation, where the main network utilized for downstream tasks is the teacher network. Moreover, the teacher network is not directly trained via gradient descent optimization; rather, it is constructed from past iterations of the student network using exponential moving average (EMA) on the student weights. This learning approach helps prevent the networks from encountering the collapse issue previously mentioned.

In specific, at each iteration, as illustrated in \cref{fig:network_arch}, Given an input image $\mathbf{I}$, both student and teacher networks output probability distributions over $K$ dimensions denoted by $\mathbf{p}_s \in \mathbb{R}^K$ and $\mathbf{p}_t \in \mathbb{R}^K$, respectively.
These probabilities are obtained by normalizing the output of the network with a softmax function. More precisely,
\begin{equation}
  \mathbf{p}_s^{(i)} = \frac{\exp(f_{\theta_s}(\mathcal{T}_s(\mathbf{I}))^{(i)} / \tau_s)}{\sum_{k=1}^K \exp(f_{\theta_s}(\mathcal{T}_s(\mathbf{I}))^{(k)} / \tau_s)},
\end{equation}   
\begin{equation}
   \mathbf{p}_t^{(i)} = \frac{\exp(f_{\theta_t}(\mathcal{T}_t(\mathbf{I}))^{(i)} / \tau_t)}{\sum_{k=1}^K \exp(f_{\theta_t}(\mathcal{T}_t(\mathbf{I}))^{(k)} / \tau_t)},
\end{equation}   
with $\tau_s, \tau_t>0$ a temperature parameter that controls the sharpness of the output distribution.
Given a fixed teacher network~$f_{\theta_t}$, we learn to match these distributions by minimizing the cross-entropy loss with respect to the parameters of the student network $\theta_s$:
\begin{equation}
        \mathcal{L}_{pretext} = \min_{\theta_s} H(\mathbf{p}_t, \mathbf{p}_s),
  \label{eq:kd}                                           
\end{equation}
where $H(a, b) = - a \log b$. Regarding teacher network's parameter $\theta_t$, it is updated it from past iterations of the student network.
Specifically, we adopt an exponential moving average (EMA) on the student weights, i.e., a momentum encoder~\cite{he2020momentum}, to update our teacher network. 
The update rule is $ \theta_t \leftarrow \lambda \theta_t + (1-\lambda) \theta_s,$ with $\lambda$ following a cosine schedule from $0.996$ to $1$ during training, followed ~\cite{grill2020bootstrap, caron2021emerging}. When the pretext task is finished training, the optimized teacher network's parameter will be used as initialized weight of the Backbone network for the downstream task.

\subsection{Downstream Task}
\label{sec:downstream}
In the downstream phase, we begin by initializing the pretrained weights of the Backbone Network into our S3Former. 
After the image is passed through the pretrained Backbone Network, we obtained the multi-scale features $\mathbf{F}^{(1)}, \mathbf{F}^{(2)}, \mathbf{F}^{(3)}, \mathbf{F}^{(4)}$. These features are now captured stronger representations of aerial imagery, trained from the pretext task. 
Subsequently, the \textbf{Pixel Decoder} module, coupled with the deformable multi-scale transformer encoder~\cite{zhu2020deformable}, enhances the relationships between the multi-scale feature maps created by the backbone and learn their correlations. This process leads to richer encoded features $\mathbf{D}^{(1)}, \mathbf{D}^{(2)}, \mathbf{D}^{(3)}, \mathbf{D}^{(4)}$.

Next, the features $\mathbf{D}^{(1)}, \mathbf{D}^{(2)}, \mathbf{D}^{(3)}, \mathbf{D}^{(4)}$ are taken as input for the \textbf{Masked Attention Mask Transformer} module to predict the segmentation mask of the solar panels. 

Firstly, the feature $\mathbf{D}^{(1)} \in \mathbb{R}^{C{\mathbf{F}^{(1)}}\times \frac{H}{4} \times \frac{W}{4}}$ is scaled to the image's original spatial shape of $H \times W$, creating the \textit{per-pixel embeddings} $\mathbf{E}_{pixel} \in \mathbb{R}^{Ce \times H \times W}$ by using a sequence of two $2\times2$ transposed convolutional layers with stride $2$. Each pixel embedding of dimension $C_e$ in $\mathbf{E}_{pixel}$ represents the segmentation feature of the corresponding pixel on the original image.

Secondly, N query embeddings $\mathbf{Q} \in \mathbb{R}^{N \times C_e}$ is proposed to represent the $N$ proposal segmentation masks of the solar panels. 
These $N$ query embeddings $\mathbf{Q}$ is further passed through a \textit{Masked Attention Transformer Decoder}~\cite{de2023solarformer}. 
This module decodes $\mathbf{Q}$ from the encoded feature maps $\mathbf{D}^{(1)}, \mathbf{D}^{(2)}, \mathbf{D}^{(3)}, \mathbf{D}^{(4)}$. The decoding procedure is done recurrently, where each step is treated as a layer and there will be 4 layers. We process each encoded feature from the lowest to the highest resolution, i.e we start from $\mathbf{D}^{(4)}$ all the way to $\mathbf{D}^{(1)}$. At each decoding layer, the query $\mathbf{Q}^{l+1}$ is decoded from the previous layer's $\mathbf{Q}^{(l)}$ and the feature map $\mathbf{D}^{4-l}$.

Furthermore, at each layer $l$, an intermediate set of predicted proposal masks $\widetilde{\mathbf{M}}_l \in \mathbb{R}^{N\times H\times W}$ is computed by correlating $\mathbf{Q}^{(l)}$ with per-pixel embeddings $\mathbf{E}_{pixel}$ via dot product across the feature space $C_e$. Mathematically, $\widetilde{\mathbf{M}}_l = \mathbf{Q}^{(l)} \otimes \mathbf{E}_{pixel}$. These predicted masks $\widetilde{\mathbf{M}}_l$ are then utilized as attention masks to highlight salient regions on the feature map $\mathbf{D}^{(4-l)}$. This selective focus aids the model in concentrating on salient features, enhancing the efficiency and convergence speed of the decoding process.

After the decoding process is finished, the last layer queries $\mathbf{Q}^{(3)}$ is passed through two multi-layer perceptrons (MLPs) $f_{mask}$ and $f_{class}$, respectively, to obtain the mask embedding $\mathbf{Q}_{mask} \in \mathbb{R}^{N\times C_e}$ and the classification embedding $\mathbf{C} \in \mathbb{R}^{N}$.

Subsequently,
the model generates the final proposal segmentation masks $\hat{\mathbf{M}} \in \mathbb{R}^{N \times  H \times W}$ as $\hat{\mathbf{M}} = \mathbf{Q}_{mask} \otimes \mathbf{E}_{pixel}$.
\begin{subequations}
\begin{align}
    \mathbf{Q}_{mask} &= f_{mask}(\mathbf{Q}^{(3)})\\
    \hat{\mathbf{M}} &= \mathbf{Q}_{mask} \otimes \mathbf{E}_{pixel} \\
    \mathbf{C} &= f_{class}(\mathbf{Q}^{(3)})
\end{align}
\end{subequations}

\textbf{Loss function}.
Let $\mathbf{M}^{gt} \in \mathbb{R}^{H\times W}$ denote the ground truth segmentation mask. The training procedure at each iteration for the downstream task optimizes the loss function $\mathcal{L}_{downstream}$ as follows: first, it identifies the index $\hat{i}$ of the proposal segmentation mask that minimizes the cross-entropy loss with the ground truth mask, as shown in \cref{eq:matching}. Subsequently, gradient descent optimization is performed on the downstream loss $\mathcal{L}_{downstream}$, which represents the cross-entropy loss between $\hat{M}_{\hat{i}}$ and $\mathbf{M}^{gt}$ (refer to \cref{eq:downstream_loss}).
\begin{equation}
       \hat{i} = \arg\min_{i \in {1..N}} \Big[ -\log(\mathbf{C}_i) +  \mathcal{L}_{CE} (\mathbf{M}^{gt}, \hat{\mathbf{M}}_i) \Big]
\label{eq:matching}
\end{equation}

\begin{equation}
    \mathcal{L}_{downstream} =-\log(\mathbf{C}_{\hat{i}}) + \mathcal{L}_{CE} (\mathbf{M}^{gt}, \hat{\mathbf{M}}_{\hat{i}})
\label{eq:downstream_loss}
\end{equation}

\textbf{Inference} At test time, all $N$ proposal masks  $\hat{\mathbf{M}}$ is joined together based on classification embedding $\mathbf{C}$ into one final mask $\mathbf{M} \in \mathbb{R}^{H\times W}$ by compute the dot product between the classification embedding $C$ and the predicted proposals mask $\hat{\mathbf{M}}$ as follow:
\begin{equation}
    \mathbf{M} = \mathbf{C}^{T}\otimes \hat{\mathbf{M}}
\end{equation}

\begin{table*}
\renewcommand{\arraystretch}{1.0}
\centering
\caption{Datasets characteristics comparison. "Positive Samples" indicate images containing solar panels, while "Negative Samples" lack solar arrays.}
\resizebox{\linewidth}{!}{
\begin{tabular}{l|cc|cc|cc}
\toprule
 \multirow{2}{*}{\textbf{ Fold }} & \multicolumn{6}{c}{\textbf{Solar PV Datasets division}} \\ \cline{2-7}
  &  \multicolumn{2}{c|}{GGE, France} & \multicolumn{2}{c|}{USGS, California} & \multicolumn{2}{c}{IGN, France}  \\ \cline{2-7}
  &  \textbf{Negative Samples} & \textbf{Positive Samples} & \textbf{Negative Samples} & \textbf{Positive Samples} & \textbf{Negative Samples} & \textbf{Positive Samples}\\ \hline
Train & 9,312 & 7,968 & 11,100 & 11,496 & 5,784 & 4608 \\
Test  & 3,104 & 2,656 & 3,700 & 3,832 & 1,928 & 1,536 \\
Validation & 3,104 & 2,656 & 3,700 & 3,832 & 1,928 & 1,536 \\
\hline
Total  & \multicolumn{2}{c|}{28,807} & \multicolumn{2}{c|}{37,660} & \multicolumn{2}{c}{17,325} \\
 \bottomrule
\end{tabular}
}
\label{tab:comparison_sample_folds}
\end{table*}

\section{Experiments}

\subsection{Datasets}

Our S3Former is benchmarked on three RGB aerial imagery datasets. We split them into 60/20/20 for train, test, and validation, dividing PV installation images equally among folds (Table \ref{tab:comparison_sample_folds}). The datasets, each containing PV installations and background, were gathered from:

\begin{itemize}
    \item \textbf{USGS, California} \cite{California2016}: 601 TIF images ($5000\times 5000$, $30$ cm/pixel), cropped to $400\times400$, totaling 37660 images (50.87\% with PVs).
    \item \textbf{IGN, France} \cite{Kasmi_2023}: 17325 thumbnails ($400\times400$, 20 cm/pixel), with 44.34\% containing solar installations.
    \item \textbf{USGS, France} \cite{Kasmi_2023}: 28807 thumbnails ($400\times400$, 10 cm/pixel), with 46.11\% showing solar installations.
\end{itemize}

\begin{table*}[!t]
\renewcommand{\arraystretch}{1.0}
\centering
\caption{Performance comparison between our S3Former with existing \textbf{DL-based Solar PV profiling methods}. The \textbf{bold} and \underline{\textit{ITALIC UNDERLINE}} represent the best and second best performances.}
\setlength{\tabcolsep}{7pt}
\begin{tabular}{l|ccc|ccc|ccc}
\toprule
 \multirow{3}{*}{\textbf{ Methods }} & \multicolumn{9}{c}{\textbf{Solar PV Segmentation Performance on each dataset}} \\ \cline{2-10}
  &  \multicolumn{3}{c|}{GGE, France} & \multicolumn{3}{c|}{USGS, California} & \multicolumn{3}{c}{IGN, France}  \\ \cline{2-10}
  &  \textbf{IoU} & \textbf{F1-score} & \textbf{Accuracy} & \textbf{IoU} & \textbf{F1-score} & \textbf{Accuracy} & \textbf{IoU} & \textbf{F1-score} & \textbf{Accuracy}\\ \hline
Zech et al.,~\cite{9300636} & $68.59$ & $81.40$ & $77.79$ & $69.80$ & $73.29$  & $82.20$ & $38.60$ & $55.69$  & $45.19$ \\
3D-PV-Locator~\cite{MAYER2022118469} & \underline{\textit{80.70}} & \textbf{89.30} & \underline{\textit{90.70}} & $80.60$ & $89.30$ & $87.40$ & $53.10$ & $69.40$  & $66.40$ \\
Zhuang, et al.,~\cite{ZHUANG2020106283}& $76.60$ & $86.69$  & $84.20$ & \underline{\textit{84.39}} & \underline{\textit{91.60}}  & \underline{\textit{90.89}} & $48.50$ & $65.30$  & $59.60$\\
Hyperion-Solar-Net   & \textbf{81.49} & $88.46$ & $89.80$ & $83.04$ & $90.73$ & $90.74$ & \underline{\textit{58.80}} & \underline{\textit{74.06}} & \underline{\textit{81.67}}  \\\hline
\textbf{Our S3Former}  & $79.56$ & \underline{\textit{88.61}} & \textbf{93.20} & \textbf{89.05} & \textbf{94.21} & \textbf{94.33} & \textbf{59.22} & \textbf{74.39} & \textbf{82.96}\\

 \bottomrule
\end{tabular}
\label{tab:comparison_solarpv}
\end{table*}

\begin{table*}[!t]
\renewcommand{\arraystretch}{1.0}
\centering
\caption{Performance comparison between our S3Former with existing \textbf{DL-segmentation Networks} on various \textbf{backbones}. The \textbf{bold} and \underline{\textit{ITALIC UNDERLINE}} represent the best and second best performances for each backbone network}
\setlength{\tabcolsep}{6pt}
\begin{tabular}{l|l|ccc|ccc|ccc}
\toprule
\multirow{3}{*}{\textbf{Backbones }} & \multirow{3}{*}{\textbf{ Networks }} & \multicolumn{9}{c}{\textbf{Solar PV Segmentation Performance on each dataset}} \\ \cline{3-11}
 &  &  \multicolumn{3}{c|}{GGE, France} & \multicolumn{3}{c|}{USGS, California} & \multicolumn{3}{c}{IGN, France}  \\ \cline{3-11}
 &  &  \textbf{IoU} & \textbf{F1-score} & \textbf{Accuracy} & \textbf{IoU} & \textbf{F1-score} & \textbf{Accuracy} & \textbf{IoU} & \textbf{F1-score} & \textbf{Accuracy}\\ 
\toprule
\multirow{7}{*}{ResNet-50}  & FCN  \cite{long2015fully} & $74.10$ & $85.12$ & $86.59$ & $63.02$ & $77.32$ & $75.55$ & $45.23$ & $62.29$  & $59.97$\\
  & UNet \cite{ronneberger2015u} & $76.60$ & $86.69$  & $84.20$ & $84.39$ & $91.60$  & $90.89$ & $48.50$ & $65.30$  & $59.60$\\
  & PSPNet \cite{zhao2017pyramid} & $77.79$ & $87.50$ & $85.50$ & $77.30$ & $87.19$ & $86.10$ & $50.90$ & $67.50$  & $62.90$ \\
  & UperNet \cite{xiao2018unified} & \underline{\textit{79.40}} & \underline{\textit{88.49}} & $89.80$ & $84.50$ & $91.60$ & $90.49$ & $52.89$ & $69.19$  & $65.49$ \\
  & Mask2Former \cite{cheng2022masked} & $74.14$ & $85.15$ & \underline{\textit{90.73}} & \underline{\textit{85.33}} & \underline{\textit{92.08}} & \underline{\textit{92.80}} & \underline{\textit{54.09}} & \underline{\textit{70.21}} &  \textbf{87.63}\\ \cline{2-11}
  & \textbf{Our S3Former}  & \textbf{79.56} & \textbf{88.61} & \textbf{93.20} & \textbf{89.05} & \textbf{94.21} & \textbf{94.33} & \textbf{59.22} & \textbf{74.39} & \underline{\textit{82.96}}\\

\midrule
\multirow{7}{*}{ResNet-101}  & FCN  \cite{long2015fully} & $73.20$ & $84.53$ & $87.14$ & $61.83$ & $76.41$ & $73.55$ & $45.52$ & $62.55$ & $62.43$ \\
 & UNet \cite{ronneberger2015u} & $68.59$ & $81.40$ & $77.79$ & $69.80$ & $73.29$  & $82.20$ & $38.60$ & $55.69$  & $45.19$\\
 & PSPNet \cite{zhao2017pyramid} & $78.29$ & $86.29$ & $87.80$ & $76.70$ & $86.79$ & $85.50$ & $48.80$ & $65.60$ & $59.10$\\
 & UperNet \cite{xiao2018unified} & \underline{\textit{79.19}} & \underline{\textit{88.40}} & $90.10$ & $83.79$ & $89.30$ & $91.20$ & \underline{\textit{52.20}} & \underline{\textit{68.59}}  & $65.10$\\

& Mask2Former \cite{cheng2022masked} & $77.03$ & $87.02$ & \underline{\textit{92.39}} & \underline{\textit{86.98}} & \underline{\textit{93.03}} & \underline{\textit{94.10}} & $49.34$ & $66.08$  & \underline{\textit{66.52}}\\
\cline{2-11}
& \textbf{Our S3Former}  & \textbf{79.68} & \textbf{88.69} & \textbf{92.91} & \textbf{89.24} & \textbf{94.13} & \textbf{94.56} & \textbf{58.69} & \textbf{73.89} & \textbf{82.87}\\
\bottomrule
\end{tabular}
\label{tab:compare_segmentation}
\end{table*}

\subsection{Evaluation Metrics}


To assess the model's performance, three common semantic segmentation scores are used: Intersection over Union (IoU), F1 score, and Accuracy. These metrics measure the accuracy of the predicted segmentation mask $\mathbf{M}$ against the ground truth segmentation mask $\mathbf{M}^{gt}$. Accuracy denotes pixel-wise accuracy, representing the proportion of correctly classified pixels over the total. IoU measures mask similarity by computing the intersection over the union of the two masks:
\begin{equation}
IoU = \frac{|\mathbf{M} \cap \mathbf{M}^{gt}|}{|\mathbf{M} \cup \mathbf{M^{gt}}|}
\end{equation}

Lastly, F1-score is computed on the comprehension of four key values: true positive (TP), true negative (TN), false positive (FP), and false negative (FN). Each of them is denoted as:

\begin{subequations}
\footnotesize
\begin{align}
    \text{TP} & = \sum_{h=1}^{H}\sum_{w=1}^{W} \mathbf{M}^{gt}_{h,w} \land \mathbf{M}_{h,w}; 
    \text{TN} & = \sum_{h=1}^{H}\sum_{w=1}^{W} \lnot (\mathbf{M}^{gt}_{h,w} \lor \mathbf{M}_{h,w}) \\
    \text{FP} & = \sum_{h=1}^{H}\sum_{w=1}^{W} \lnot \mathbf{M}^{gt}_{h,w} \land \mathbf{M}_{h,w}; 
    \text{FN} & = \sum_{h=1}^{H}\sum_{w=1}^{W} \mathbf{M}^{gt}_{h,w} \land \lnot \mathbf{M}_{h,w}
\end{align}
\end{subequations}
Then, F1-score is computed as follow:
\begin{equation}
    \text{F1-score} = \frac{\text{TP}}{\text{TP} + \frac{1}{2} (\text{FP} + \text{FN})} 
\end{equation}




\subsection{Baselines}
\textbf{Solar PV profiling methods.} The S3Former was compared with SOTA solar PV profiling methods, namely Zech et al. ~\cite{9300636}, 3D-PV-Locator ~\cite{MAYER2022118469}, Zhuang, et al. ~\cite{ZHUANG2020106283} and Hyperion-Solar-Net \cite{parhar2022hyperionsolarnet}. The evaluation was conducted across three datasets using ResNet 101 as the backbone architecture.

\textbf{Deep Learning (DL) Segmentation methods.} To authors' knowledge, there are currently no solar PV profiling methods employing an end-to-end network for predicting solar panel segmentation without prior localization. Therefore, state-of-the-art DL segmentation networks are employed for comparison with the S3Former model. These methods include FCN~\cite{long2015fully}, UNet~\cite{ronneberger2015u}, PSPNet~\cite{zhao2017pyramid}, UperNet~\cite{xiao2018unified}, Mask2Former~\cite{cheng2022masked}. The evaluation is performed across three datasets, utilizing ResNet-50 and ResNet-101 as backbone architectures.

\subsection{Quantitative Results and Analysis}
Table \ref{tab:comparison_solarpv} summarizes the performance of S3Former against current \textbf{SOTA solar PV profiling methods}. 
This segmentation model outperforms Zech et al. ~\cite{9300636}, 3D-PV-Locator ~\cite{MAYER2022118469}, Zhuang, et al. ~\cite{ZHUANG2020106283} and Hyperion-Solar-Net \cite{parhar2022hyperionsolarnet} on all three datasets. For instance, when analyzing the USGS California dataset, the S3Former model exhibits substantial improvements, including an impressive 4.66\% increase in IoU, a notable 2.61\% enhancement in F1-score, and a substantial boost of 3.56\% in Accuracy compared to the second-best CNN method Zhuang, et al.,~\cite{ZHUANG2020106283}. Similarly, on IGN France, the model showcases remarkable advancements, with gains of 0.42\%, 0.33\%, and 1.29\% in IoU, F1-score, and Accuracy, respectively, when compared to the runner-up method, Hyperion-Solar-Net \cite{parhar2022hyperionsolarnet}.


\begin{table*}[]
\centering
\caption{Ablation study on augmentations use in pretext tasks for our S3Former, employing ResNet-50 Backbone across three datasets.}
\resizebox{\textwidth}{!}{ 
\begin{tabular}{ccc|ccc|ccc|ccc}
\toprule
 \multicolumn{3}{c|}{\textbf{Augmentations}} &  \multicolumn{9}{c}{\textbf{Segmentation performance on each datasets}} \\ \hline
     & & & \multicolumn{3}{c|}{GGE, France} & \multicolumn{3}{c|}{USGS, California} & \multicolumn{3}{c}{IGN, France}  \\ \cline{4-12}
    \shortstack{\textbf{Color}\\\textbf{Jitter}}& \shortstack{\textbf{Gaussian}\\\textbf{Noise}}& \shortstack{\textbf{Horizontal}\\\textbf{Flip}} &\textbf{IoU} & \textbf{F1-score} & \textbf{Accuracy} & \textbf{IoU} & \textbf{F1-score} & \textbf{Accuracy} & \textbf{IoU} & \textbf{F1-score} & \textbf{Accuracy}\\ 
\hline
  \xmark & \xmark &  \xmark & $75.99$& $86.25$ & $92.20$	& $86.28$	& $92.56$	& $92.76$	& $57.35$	&$72.74$	& $77.53$ \\ \cline{1-12}
  \xmark & \xmark &  \cmark & $76.43$ & $86.64$ & $92.61$ & $86.72$ & $92.89$ & $93.13$ & $57.77$ & $73.23$ & $77.87$ \\ 
 \cline{1-12}
   \xmark & \cmark & \xmark & $79.36$ & $88.49$ & $90.99$ & $86.38$ & $92.69$  & $93.59$ & $54.40$ & $70.46$ & $83.94$ \\ \cline{1-12}
   \cmark & \xmark & \xmark & $79.34$ & $88.46$ & $92.65$ & $88.00$ & $93.62$ & $92.95$  & $56.93$ & $72.56$ & $80.35$ \\ \cline{1-12}
  \cmark & \cmark &  \cmark & $79.56$ & $88.61$  & $93.20$ & $89.05$ & $94.21$ & $94.33$ & $59.22$ & $74.39$ & $82.96$ \\             
\bottomrule
\end{tabular}
}
\label{tab:ablation}
\end{table*}

\begin{figure*}[!ht]
\centering
\includegraphics[width=\linewidth]{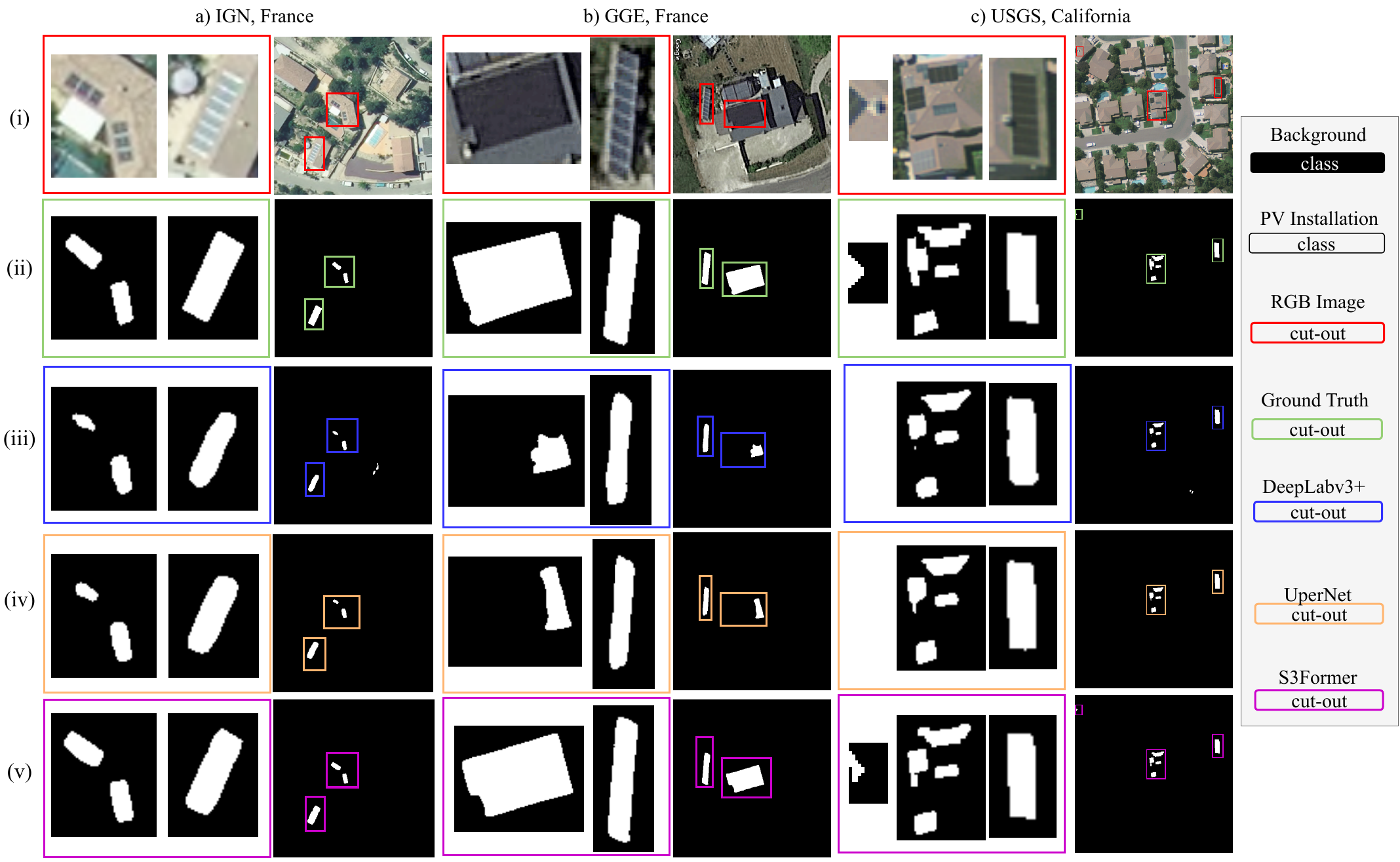}
\caption{Qualitative comparison on (a) IGN France, (b) GGE France, (c) USGS California. From top to bottom: (i) Original RGB Image, (ii) Groundtruth, (iii) Upernet (\cite{xiao2018unified}), (iv) DeepLabv3+ (\cite{chen2017rethinking}) and (v) S3Former.}
\label{fig:solar_pv_comparison}
\end{figure*}

Table \ref{tab:compare_segmentation} presents performance comparisons between S3Former and the SOTA \textbf{DL segmentation methods}. Overall, S3Former achieves highest performance on most datasets and metrics across different backbone  ResNet-50 and ResNet-101. Specifically, regarding \textit{GGE, France.}, S3Former demonstrate significant improvements over most of SOTA methods.  
On the ResNet-50 backbone, S3Former outperforms the second best method, UperNet (~\cite{cheng2022masked}), by 0.16\% on IoU and 0.12\% on F1-score and improves Mask2former(~\cite{cheng2022masked}) by 2.47\% on Accuracy.
When benchmarked on the ResNet-101 backbone, 
S3Former gains 0.49\% IoU and 0.29\% F1-score on UperNet(~\cite{xiao2018unified}) and improves the Accuracy result of Mask2former(~\cite{cheng2022masked}) by 0.52\%. On \textit{USGS, California.}, S3Former displays notable gains over SOTA methods on both ResNet backbones. On the ResNet-101 backbone, when compared to the top performing CNN-based model UperNet(~\cite{xiao2018unified}). The difference with Mask2former(~\cite{cheng2022masked}) is still remarkable at 1.8\%, 1.02\% and 0.29\%. For ResNet-50, S3Former improves 3.10\% on Accuracy with respect to UperNet(~\cite{xiao2018unified}). 
On \textit{IGN, France.}, despite its lower resolution, S3Former exhibits robust scores against other models on the ResNet backbones. The dataset's relatively modest performance can be attributed to its lower resolution and limited availability of imagery. Similar to the GGE France dataset, the granularity in the imagery could impact the model's efficiency in detecting installations. Notably, when benchmarked on the ResNet-101 backbone, S3Former showcases improvements of 4.21\% on IoU, 2.54\% on UperNet (\cite{xiao2018unified}), and 13.14\% on Accuracy in comparison to Mask2Former (\cite{cheng2022masked}). On the ResNet-50 backbone, S3Former improves the results of Mask2Former (\cite{cheng2022masked}) by 5.13\% and 4.18\% on IoU and F1-score, respectively.

\begin{figure*}[!t]
\centering
\includegraphics[width=.82\linewidth]{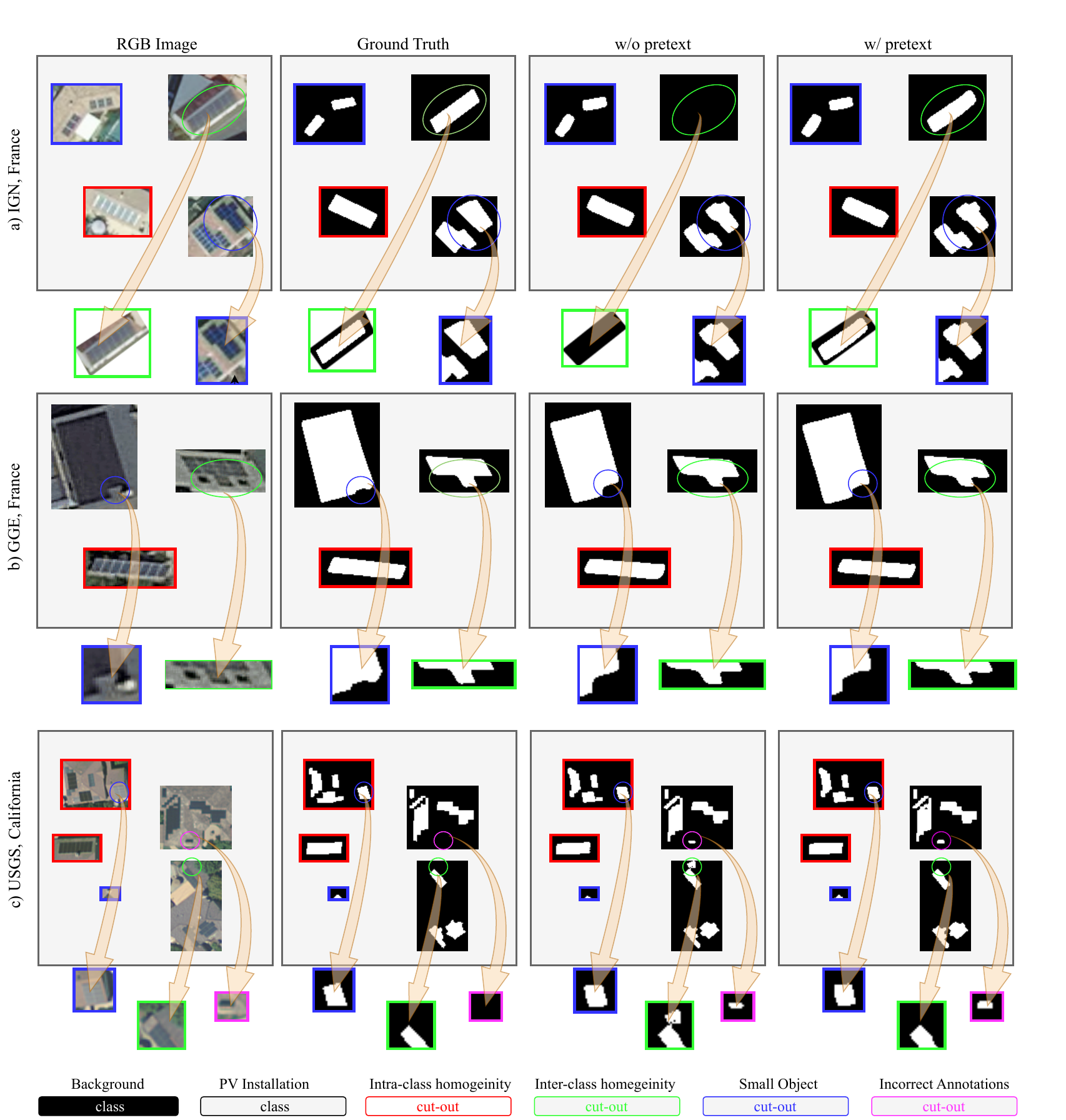}
\caption{Extended qualitative comparison on different datasets (a) IGN France, (b) GGE France, (c) USGS California. From left to right: RGB Image, Ground-Truth, w/o pretext and w/ pretext. Special cases highlighting the strength of both models and improvements of S3Former with respect to pretext task pretraining were selected: inter-class homogeneity (green), intra-class heterogeneity (red), and small-object identification (blue). Cases of missing annotations on the data were highlighted (purple). }
\label{fig:SF-VS-S3F}
\end{figure*}

\subsection{Ablation study}

\indent
Table \ref{tab:ablation} illustrates the difference between the S3Former network performance using different augmentations, namely Color Jitter, Gaussian Noise and Horizontal Flip. The baseline performance without any augmentations exhibits decent results across all datasets, with the highest scores observed in GGE, France. Introducing flip augmentation marginally improves performance, while Gaussian augmentation notably enhances it, particularly benefiting the GGE, France and USGS, California datasets.
Moreover, color augmentation leads to improved performance, especially noticeable in the IGN, France dataset, though its impact is less pronounced compared to Gaussian augmentation. The combination of all augmentations yields the best overall performance across all datasets, indicating their collective positive contribution to the model's ability to generalize across diverse datasets.

\subsection{Qualitative Results and Analysis}

\textbf{Comparison between S3Former and DL-Segmentation Networks}
\indent
Fig. \ref{fig:solar_pv_comparison} provides a qualitative comparison of our S3Former against widely-used baseline models: DeepLabV3+ (\cite{chen2017rethinking}) and UperNet (\cite{xiao2018unified}). We highlight the advancements of our model in addressing challenges previously outlined on:

\noindent
\underline{\textit{Small objects}}: 
Fig. \ref{fig:solar_pv_comparison} highlight S3Former's proficiency in identifying small objects, even as diminutive as solar cells measuring $10 \times 5$ pixels, underscoring its potential in high-resolution aerial imagery. Specifically, in the USGS, California example (Fig. \ref{fig:solar_pv_comparison}c)), S3Former excels in identifying a small PV module situated at the top-left of the image, a task where the compared methods fail. Fig. \ref{fig:solar_pv_comparison} a) further reinforces this observation by illustrating the enhancements made by S3Former in accurately segmenting the target PV installations.

\noindent
\underline{\textit{Intra-class heterogeneity}}:
Figs. \ref{fig:solar_pv_comparison} a), b), c) illustrate the diverse PV installations even within a single rooftop. The variety in shapes, colors, and orientations poses challenges for deep learning. Notably, S3Former accurately maps varied color cells in Fig. \ref{fig:solar_pv_comparison} a) and identifies accurate boundary in Fig. \ref{fig:solar_pv_comparison} b).

\noindent
\underline{\textit{Inter-class heterogeneity}}: Figs. \ref{fig:solar_pv_comparison} b), c) depict challenges due to visual similarities between PV installations and other elements. In Fig. \ref{fig:solar_pv_comparison} b), while both DeepLabV3+ and UperNet falter due to the rooftop's similar tone to PV installations, S3Former accurately masks it. In Fig. \ref{fig:solar_pv_comparison} c), DeepLabV3+ erroneously misidentifies a pool reflection as a solar cell. Pools and windows are prone to create reflections under certain solar conditions which, due its similar tone and shape with PV modules, are mistaken by segmentation networks.

\noindent
\underline{\textit{Overall performance}}: Fig. \ref{fig:solar_pv_comparison} exhibits S3Former's superior ability to accurately depict PV installations, outperforming baseline models in clarity and precision. It consistently identifies minute panels throughout the image and consistently offers robust results with minimal background confusion.

\textbf{Comparison between S3Former with and without pretext task}
In this section, we conduct a qualitative comparison between S3Former with and without pretext task pretraining to validate the enhancements of it. We will revisit the challenges discussed in the previous section, as highlighted in Fig. \ref{fig:SF-VS-S3F}.

\noindent
\underline{\textit{Small objects}}: 
Fig. \ref{fig:SF-VS-S3F} showcases the improvements of S3Former with pretext task pretraining. Fig. \ref{fig:SF-VS-S3F} a) and c) showcase sharper lines and more accurate boundaries on solar modules produced by S3Former w/ pretext task. This behavior is reaffirmed in Panel b), where S3Former w/ pretext task effectively masks a small indent at the bottom of the solar installation.


\noindent
\underline{\textit{Inter-class heterogeneity}}: Figs. \ref{fig:SF-VS-S3F} b), depicts a case where CNN models traditionally struggle: an object situated next to a solar installation with a similar size and color. Both models accurately discard this element as part of the solar mask. However, Figs. \ref{fig:SF-VS-S3F} a) and c) contain instances of S3Former w/ pretext task outperforming without pretraining with pretext task in identifying erroneous elements and eliminating them from the result mask.

\noindent
\underline{\textit{Overall performance}}: Fig. \ref{fig:SF-VS-S3F}  exhibits S3Former's superior capability to delineate small installations and enhance the model's proficiency in distinguishing solar cells from background objects. This reinforces the robust features extracted by the pretext task pretraining in S3Former.

\section{Conclusion}



We present S3Former, a novel approach aimed at segmenting solar panels from aerial imagery and providing crucial size and location data essential for grid impact analysis of such installations. S3Former features a Masked Attention Mask Transformer coupled with a self-supervised learning pretrained backbone. Our model effectively utilizes both low-level and high-level features extracted from the backbone and incorporates an instance query mechanism within the Transformer architecture to enhance solar PV installation localization. We introduce a self-supervised learning (SSL) phase (pretext task) to enhance the initialization weights on the backbone of S3Former, resulting in noticeable improvements in performance. Through rigorous evaluation using diverse datasets from GGE (France), IGN (France), and USGS (California, USA) across various GSDs, our extensive experiments consistently demonstrate that S3Former either matches or surpasses state-of-the-art models (SOTA).


\footnotesize
\bibliographystyle{IEEEtran}
\bibliography{journal_references}

\begin{thebibliography}{10}
\providecommand{\url}[1]{#1}
\csname url@samestyle\endcsname
\providecommand{\newblock}{\relax}
\providecommand{\bibinfo}[2]{#2}
\providecommand{\BIBentrySTDinterwordspacing}{\spaceskip=0pt\relax}
\providecommand{\BIBentryALTinterwordstretchfactor}{4}
\providecommand{\BIBentryALTinterwordspacing}{\spaceskip=\fontdimen2\font plus
\BIBentryALTinterwordstretchfactor\fontdimen3\font minus \fontdimen4\font\relax}
\providecommand{\BIBforeignlanguage}[2]{{%
\expandafter\ifx\csname l@#1\endcsname\relax
\typeout{** WARNING: IEEEtran.bst: No hyphenation pattern has been}%
\typeout{** loaded for the language `#1'. Using the pattern for}%
\typeout{** the default language instead.}%
\else
\language=\csname l@#1\endcsname
\fi
#2}}
\providecommand{\BIBdecl}{\relax}
\BIBdecl

\bibitem{PETERS20192732}
I.~M. Peters, C.~D.~R. Gallegos, S.~E. Sofia, and T.~Buonassisi, ``The value of efficiency in photovoltaics,'' \emph{Joule}, vol.~3, no.~11, pp. 2732--2747, 2019.

\bibitem{LEMAY2023113571}
A.~C. Lemay, S.~Wagner, and B.~P. Rand, ``Current status and future potential of rooftop solar adoption in the united states,'' \emph{Energy Policy}, vol. 177, p. 113571, 2023.

\bibitem{EIA2022}
``Record u.s. small-scale solar capacity was added in 2022,'' \url{https://www.eia.gov/todayinenergy/detail.php?id=603411}.

\bibitem{PIERRO2022983}
M.~Pierro, D.~Gentili, F.~R. Liolli, C.~Cornaro, D.~Moser, A.~Betti, M.~Moschella, E.~Collino, D.~Ronzio, and D.~van Der~Meer, ``Progress in regional pv power forecasting: A sensitivity analysis on the italian case study,'' \emph{Renewable Energy}, vol. 189, pp. 983--996, 2022.

\bibitem{7347457}
H.~Shaker, H.~Zareipour, and D.~Wood, ``A data-driven approach for estimating the power generation of invisible solar sites,'' \emph{IEEE Transactions on Smart Grid}, vol.~7, no.~5, pp. 2466--2476, 2015.

\bibitem{YU20182605}
J.~Yu, Z.~Wang, A.~Majumdar, and R.~Rajagopal, ``Deepsolar: A machine learning framework to efficiently construct a solar deployment database in the united states,'' \emph{Joule}, vol.~2, no.~12, pp. 2605--2617, 2018.

\bibitem{ONEILL2013413}
S.~J. O’neill, M.~Boykoff, S.~Niemeyer, and S.~A. Day, ``On the use of imagery for climate change engagement,'' \emph{Global environmental change}, vol.~23, no.~2, pp. 413--421, 2013.

\bibitem{DBLP:journals/corr/abs-1901-04277}
N.~Said, K.~Ahmad, M.~Riegler, K.~Pogorelov, L.~Hassan, N.~Ahmad, and N.~Conci, ``Natural disasters detection in social media and satellite imagery: a survey,'' \emph{Multimedia Tools and Applications}, vol.~78, pp. 31\,267--31\,302, 2019.

\bibitem{9324098}
S.~P. B, P.~Sheladiya~Kaushik, J.~Patel, C.~R. Patel, and R.~M. Tailor, ``Assessing land suitability for managing urban growth: An application of gis and rs,'' in \emph{IGARSS 2020 - 2020 IEEE International Geoscience and Remote Sensing Symposium}, 2020, pp. 4243--4246.

\bibitem{Effat2013AMA}
H.~A. Effat and M.~N. Hegazy, ``A multidisciplinary approach to mapping potential urban development zones in sinai peninsula, egypt using remote sensing and gis,'' \emph{J.of Geographic Information System}, vol. 2013, 2013.

\bibitem{NEURIPS_DATASETS_AND_BENCHMARKS2021_4e732ced}
J.~Wang, Z.~Zheng, A.~Ma, X.~Lu, and Y.~Zhong, ``Loveda: A remote sensing land-cover dataset for domain adaptive semantic segmentation,'' \emph{arXiv preprint arXiv:2110.08733}, 2021.

\bibitem{kasmi2023unsupervised}
G.~Kasmi, L.~Dubus, P.~Blanc, and Y.-M. Saint-Drenan, ``Towards unsupervised assessment with open-source data of the accuracy of deep learning-based distributed pv mapping,'' in \emph{Workshop on Machine Learning for Earth Observation (MACLEAN)}, 2022.

\bibitem{parhar2022hyperionsolarnet}
P.~Parhar, R.~Sawasaki, A.~Todeschini, H.~Vahabi, N.~Nusaputra, and F.~Vergara, ``Hyperionsolarnet: solar panel detection from aerial images,'' \emph{arXiv preprint arXiv:2201.02107}, 2022.

\bibitem{California2016}
K.~Bradbury, R.~Saboo, T.~L~Johnson, J.~M. Malof, A.~Devarajan, W.~Zhang, L.~M~Collins, and R.~G~Newell, ``Distributed solar photovoltaic array location and extent dataset for remote sensing object identification,'' \emph{Scientific data}, vol.~3, no.~1, pp. 1--9, 2016.

\bibitem{Kasmi_2023}
G.~Kasmi, Y.-M. Saint-Drenan, D.~Trebosc, R.~Jolivet, J.~Leloux, B.~Sarr, and L.~Dubus, ``A crowdsourced dataset of aerial images with annotated solar photovoltaic arrays and installation metadata,'' \emph{Scientific Data}, vol.~10, no.~1, p.~59, 2023.

\bibitem{MALOF2016229}
J.~M. Malof, K.~Bradbury, L.~M. Collins, and R.~G. Newell, ``Automatic detection of solar photovoltaic arrays in high resolution aerial imagery,'' \emph{Applied energy}, vol. 183, pp. 229--240, 2016.

\bibitem{8127092}
J.~M. Malof, L.~M. Collins, and K.~Bradbury, ``A deep convolutional neural network, with pre-training, for solar photovoltaic array detection in aerial imagery,'' in \emph{2017 IEEE International Geoscience and Remote Sensing Symposium (IGARSS)}, 2017, pp. 874--877.

\bibitem{ZHUANG2020106283}
L.~Zhuang, Z.~Zhang, and L.~Wang, ``The automatic segmentation of residential solar panels based on satellite images: A cross learning driven u-net method,'' \emph{Applied Soft Computing}, vol.~92, p. 106283, 2020.

\bibitem{9300636}
M.~Zech and J.~Ranalli, ``Predicting pv areas in aerial images with deep learning,'' in \emph{2020 47th IEEE Photovoltaic Specialists Conference (PVSC)}, 2020, pp. 0767--0774.

\bibitem{MAYER2022118469}
K.~Mayer, B.~Rausch, M.-L. Arlt, G.~Gust, Z.~Wang, D.~Neumann, and R.~Rajagopal, ``3d-pv-locator: Large-scale detection of rooftop-mounted photovoltaic systems in 3d,'' \emph{Applied Energy}, vol. 310, p. 118469, 2022.

\bibitem{vaswani2017attention}
A.~Vaswani, N.~Shazeer, N.~Parmar, J.~Uszkoreit, L.~Jones, A.~N. Gomez, L.~u. Kaiser, and I.~Polosukhin, ``Attention is all you need,'' in \emph{Advances in Neural Information Processing Systems}, I.~Guyon, U.~V. Luxburg, S.~Bengio, H.~Wallach, R.~Fergus, S.~Vishwanathan, and R.~Garnett, Eds., vol.~30.\hskip 1em plus 0.5em minus 0.4em\relax Curran Associates, Inc., 2017.

\bibitem{dosovitskiy2020image}
A.~Dosovitskiy, L.~Beyer, A.~Kolesnikov, D.~Weissenborn, X.~Zhai, T.~Unterthiner, M.~Dehghani, M.~Minderer, G.~Heigold, S.~Gelly \emph{et~al.}, ``An image is worth 16x16 words: Transformers for image recognition at scale,'' \emph{International Conference on Learning Representations (ICLR)}, 2020.

\bibitem{9222960}
A.~Gillioz, J.~Casas, E.~Mugellini, and O.~A. Khaled, ``Overview of the transformer-based models for nlp tasks,'' in \emph{15th Conference on Computer Science and Information Systems}, 2020, pp. 179--183.

\bibitem{touvron2021training}
H.~Touvron, M.~Cord, M.~Douze, F.~Massa, A.~Sablayrolles, and H.~J{\'e}gou, ``Training data-efficient image transformers \& distillation through attention,'' in \emph{International conference on machine learning}.\hskip 1em plus 0.5em minus 0.4em\relax PMLR, 2021, pp. 10\,347--10\,357.

\bibitem{yamazaki2022vlcap}
K.~Yamazaki, S.~Truong, K.~Vo, M.~Kidd, C.~Rainwater, K.~Luu, and N.~Le, ``Vlcap: Vision-language with contrastive learning for coherent video paragraph captioning,'' in \emph{2022 IEEE International Conference on Image Processing (ICIP)}.\hskip 1em plus 0.5em minus 0.4em\relax IEEE, 2022, pp. 3656--3661.

\bibitem{yamazaki2023vltint}
K.~Yamazaki, K.~Vo, Q.~S. Truong, B.~Raj, and N.~Le, ``Vltint: visual-linguistic transformer-in-transformer for coherent video paragraph captioning,'' in \emph{AAAI}, vol.~37, no.~3, 2023, pp. 3081--3090.

\bibitem{vo2022contextual}
K.~Vo, K.~Yamazaki, P.~X. Nguyen, P.~Nguyen, K.~Luu, and N.~Le, ``Contextual explainable video representation: Human perception-based understanding,'' in \emph{2022 56th Asilomar Conference on Signals, Systems, and Computers}.\hskip 1em plus 0.5em minus 0.4em\relax IEEE, 2022, pp. 1326--1333.

\bibitem{vo2023aoe}
K.~Vo, S.~Truong, K.~Yamazaki, B.~Raj, M.-T. Tran, and N.~Le, ``Aoe-net: Entities interactions modeling with adaptive attention mechanism for temporal action proposals generation,'' \emph{International Journal of Computer Vision}, vol. 131, no.~1, pp. 302--323, 2023.

\bibitem{sun2021sparse}
P.~Sun, R.~Zhang, Y.~Jiang, T.~Kong, C.~Xu, W.~Zhan, M.~Tomizuka, L.~Li, Z.~Yuan, C.~Wang \emph{et~al.}, ``Sparse r-cnn: End-to-end object detection with learnable proposals,'' in \emph{Proceedings of the IEEE/CVF conference on computer vision and pattern recognition}, 2021, pp. 14\,454--14\,463.

\bibitem{tran2022aisformer}
M.~Tran, K.~Vo, K.~Yamazaki, A.~Fernandes, M.~Kidd, and N.~Le, ``Aisformer: Amodal instance segmentation with transformer,'' \emph{BMVC}, 2022.

\bibitem{nguyen2023embryosformer}
T.-P. Nguyen, T.-T. Pham, T.~Nguyen, H.~Le, D.~Nguyen, H.~Lam, P.~Nguyen, J.~Fowler, M.-T. Tran, and N.~Le, ``Embryosformer: Deformable transformer and collaborative encoding-decoding for embryos stage development classification,'' in \emph{The IEEE/CVF Winter Conference on Applications of Computer Vision}, 2023, pp. 1981--1990.

\bibitem{chen2020simple}
T.~Chen, S.~Kornblith, M.~Norouzi, and G.~Hinton, ``A simple framework for contrastive learning of visual representations,'' \emph{preprint arXiv:2002.05709}, 2020.

\bibitem{dosovitskiy2016discriminative}
A.~Dosovitskiy, P.~Fischer, J.~T. Springenberg, M.~Riedmiller, and T.~Brox, ``Discriminative unsupervised feature learning with exemplar convolutional neural networks,'' \emph{TPAMI}, 2016.

\bibitem{he2020momentum}
K.~He, H.~Fan, Y.~Wu, S.~Xie, and R.~Girshick, ``Momentum contrast for unsupervised visual representation learning,'' 2020.

\bibitem{wu2018unsupervised}
Z.~Wu, Y.~Xiong, S.~X. Yu, and D.~Lin, ``Unsupervised feature learning via non-parametric instance discrimination,'' 2018.

\bibitem{gutmann2010noise}
M.~Gutmann and A.~Hyv{\"a}rinen, ``Noise-contrastive estimation: A new estimation principle for unnormalized statistical models,'' in \emph{International Conference on Artificial Intelligence and Statistics}, 2010.

\bibitem{asano2019self}
Y.~M. Asano, C.~Rupprecht, and A.~Vedaldi, ``Self-labelling via simultaneous clustering and representation learning,'' 2020.

\bibitem{caron2019unsupervised}
M.~Caron, P.~Bojanowski, J.~Mairal, and A.~Joulin, ``Unsupervised pre-training of image features on non-curated data,'' 2019.

\bibitem{grill2020bootstrap}
J.-B. Grill, F.~Strub, F.~Altch{\'e}, C.~Tallec, P.~H. Richemond, E.~Buchatskaya, C.~Doersch, B.~A. Pires, Z.~D. Guo, M.~G. Azar, B.~Piot, K.~Kavukcuoglu, R.~Munos, and M.~Valko, ``Bootstrap your own latent: A new approach to self-supervised learning,'' 2020.

\bibitem{chen2020exploring}
X.~Chen and K.~He, ``Exploring simple siamese representation learning,'' \emph{preprint arXiv:2011.10566}, 2020.

\bibitem{gidaris2020learning}
S.~Gidaris, A.~Bursuc, N.~Komodakis, P.~P{\'e}rez, and M.~Cord, ``Learning representations by predicting bags of visual words,'' 2020.

\bibitem{zbontar2021barlow}
J.~Zbontar, L.~Jing, I.~Misra, Y.~LeCun, and S.~Deny, ``Barlow twins: Self-supervised learning via redundancy reduction,'' \emph{arXiv preprint arXiv:2103.03230}, 2021.

\bibitem{caron2021emerging}
M.~Caron, H.~Touvron, I.~Misra, H.~J{\'e}gou, J.~Mairal, P.~Bojanowski, and A.~Joulin, ``Emerging properties in self-supervised vision transformers,'' in \emph{Proceedings of the IEEE/CVF international conference on computer vision}, 2021, pp. 9650--9660.

\bibitem{tarvainen2017mean}
A.~Tarvainen and H.~Valpola, ``Mean teachers are better role models: Weight-averaged consistency targets improve semi-supervised deep learning results,'' \emph{preprint arXiv:1703.01780}, 2017.

\bibitem{krizhevsky2017imagenet}
A.~Krizhevsky, I.~Sutskever, and G.~E. Hinton, ``Imagenet classification with deep convolutional neural networks,'' \emph{Communications of the ACM}, vol.~60, no.~6, pp. 84--90, 2017.

\bibitem{simonyan2014very}
K.~Simonyan and A.~Zisserman, ``Very deep convolutional networks for large-scale image recognition,'' \emph{arXiv preprint arXiv:1409.1556}, 2014.

\bibitem{he2016deep}
K.~He, X.~Zhang, S.~Ren, and J.~Sun, ``Deep residual learning for image recognition,'' in \emph{Proceedings of the IEEE conference on computer vision and pattern recognition}, 2016, pp. 770--778.

\bibitem{russakovsky2015imagenet}
O.~Russakovsky, J.~Deng, H.~Su, J.~Krause, S.~Satheesh, S.~Ma, Z.~Huang, A.~Karpathy, A.~Khosla, M.~Bernstein \emph{et~al.}, ``Imagenet large scale visual recognition challenge,'' \emph{International journal of computer vision}, vol. 115, pp. 211--252, 2015.

\bibitem{de2023solarformer}
A.~de~Luis, M.~Tran, T.~Hanyu, A.~Tran, L.~Haitao, R.~McCann, A.~Mantooth, Y.~Huang, and N.~Le, ``Solarformer: Multi-scale transformer for solar pv profiling,'' \emph{arXiv preprint arXiv:2310.20057}, 2023.

\bibitem{becker1992self}
S.~Becker and G.~E. Hinton, ``Self-organizing neural network that discovers surfaces in random-dot stereograms,'' \emph{Nature}, vol. 355, no. 6356, pp. 161--163, 1992.

\bibitem{44873}
\BIBentryALTinterwordspacing
G.~Hinton, O.~Vinyals, and J.~Dean, ``Distilling the knowledge in a neural network,'' in \emph{NIPS Deep Learning and Representation Learning Workshop}, 2015. [Online]. Available: \url{http://arxiv.org/abs/1503.02531}
\BIBentrySTDinterwordspacing

\bibitem{zhu2020deformable}
X.~Zhu, W.~Su, L.~Lu, B.~Li, X.~Wang, and J.~Dai, ``Deformable detr: Deformable transformers for end-to-end object detection,'' \emph{arXiv preprint arXiv:2010.04159}, 2020.

\bibitem{long2015fully}
J.~Long, E.~Shelhamer, and T.~Darrell, ``Fully convolutional networks for semantic segmentation,'' in \emph{Proceedings of the IEEE conference on computer vision and pattern recognition}, 2015, pp. 3431--3440.

\bibitem{ronneberger2015u}
O.~Ronneberger, P.~Fischer, and T.~Brox, ``U-net: Convolutional networks for biomedical image segmentation,'' in \emph{International Conference on Medical image computing and computer-assisted intervention}.\hskip 1em plus 0.5em minus 0.4em\relax Springer, 2015, pp. 234--241.

\bibitem{zhao2017pyramid}
H.~Zhao, J.~Shi, X.~Qi, X.~Wang, and J.~Jia, ``Pyramid scene parsing network,'' in \emph{Proceedings of the IEEE conference on computer vision and pattern recognition}, 2017, pp. 2881--2890.

\bibitem{xiao2018unified}
T.~Xiao, Y.~Liu, B.~Zhou, Y.~Jiang, and J.~Sun, ``Unified perceptual parsing for scene understanding,'' in \emph{Proceedings of the European conference on computer vision (ECCV)}, 2018, pp. 418--434.

\bibitem{cheng2022masked}
B.~Cheng, I.~Misra, A.~G. Schwing, A.~Kirillov, and R.~Girdhar, ``Masked-attention mask transformer for universal image segmentation,'' in \emph{Proceedings of the IEEE/CVF Conference on Computer Vision and Pattern Recognition}, 2022, pp. 1290--1299.

\bibitem{chen2017rethinking}
L.-C. Chen, G.~Papandreou, F.~Schroff, and H.~Adam, ``Rethinking atrous convolution for semantic image segmentation,'' \emph{arXiv preprint arXiv:1706.05587}, 2017.

\end{thebibliography}

\vfill

\end{document}